\documentclass[a4paper, 10pt, conference]{ieeeconf}      
\usepackage{FG2020}
\usepackage{graphicx}
\usepackage{booktabs}
\usepackage{balance}
\usepackage{placeins}
\usepackage{url}
\FGfinalcopy 

\IEEEoverridecommandlockouts
\def\FGPaperID{****} 

\title{\LARGE \bf
A Scalable Approach for Facial Action Unit Classifier Training Using Noisy Data for Pre-Training}

\author{
}

\begin{document}

\author{Alberto Fung$^1$, Daniel McDuff$^2$\\ $^1$University of Houston, $^2$Microsoft Research}
\maketitle

\begin{abstract}
Machine learning systems are being used to automate many types of laborious labeling tasks. Facial action coding is an example of such a labeling task that requires copious amounts of time and a beyond average level of human domain expertise. In recent years, the use of end-to-end deep neural networks has led to significant improvements in action unit recognition performance and many network architectures have been proposed. Do the more complex deep neural network (DNN) architectures perform sufficiently well to justify the additional complexity? We show that pre-training on a large diverse set of noisy data can result in even a simple CNN model improving over the current state-of-the-art DNN architectures. The average F1-score achieved with our proposed method on the DISFA dataset is 0.60, compared to a previous state-of-the-art of 0.57. Additionally, we show how the number of subjects and number of images used for pre-training impacts the model performance. The approach that we have outlined is open-source, highly scalable, and not dependent on the model architecture. We release the code and data: \url{https://github.com/facialactionpretrain/facs}. 

\end{abstract}

\section{Introduction}
Facial expressions can convey information about a person's perceived emotional state~\cite{ekman1965differential}, their intentions~\cite{fridlund2014human, horstmann2003facial,seidel2010impact}, and even physical state~\cite{pitcairn1990non}. Proper understanding of facial expressions is a vital aspect of human interaction and social communication~\cite{ekman1997face, fridlund2014human}. 
Given the significance of facial actions, there is great interest in building assistive technologies and computer systems that can leverage signals from the face.
Many of the benefits offered by facial coding are reliant on the ability to code large amounts of image or video data. 
For example, analyses of facial actions are being used to help drive increased understanding of psychology~\cite{girard2017historical} and complex medical conditions, such as psychosis (e.g., schizophrenia)~\cite{vijay2016computational} and depression~\cite{stratou2013automatic}, and can even provide a means for objective measurement of pain~\cite{kaltwang2012continuous}. In each of these cases, individual differences and contextual information adds a lot of variability to what is displayed on the face.  
Relying on manual coding severely limits how effectively research can be translated into practice as the signal-to-noise ratio is often small. Whereas, with large-scale analyses, significant trends can be observed, even in the presence of noise~\cite{girard2017historical,mcduff2019democratizing}.

\begin{figure}
    \centering
    \includegraphics[width=8.5cm]{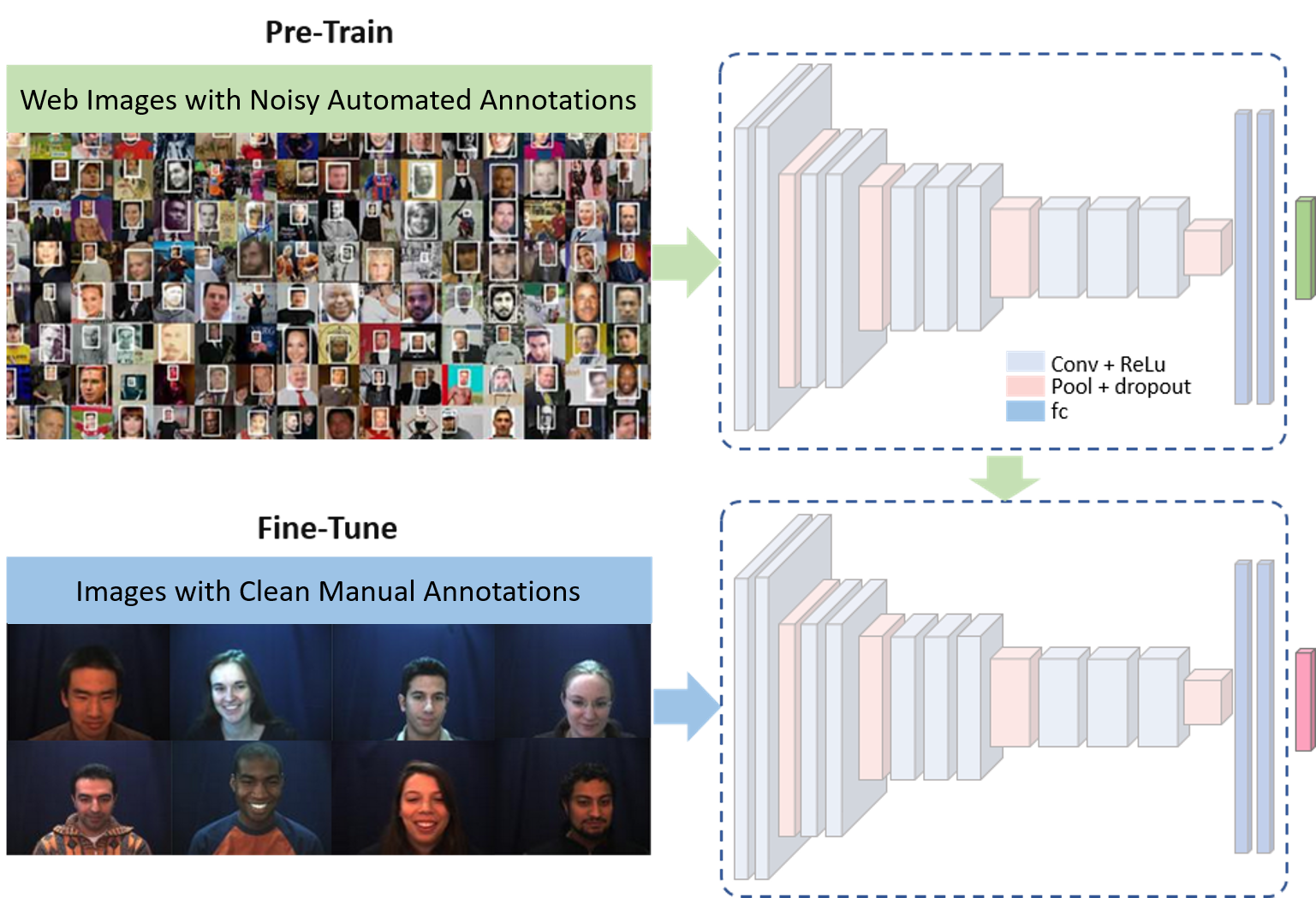}
    \caption{An overview of our approach using automatic annotation to generate a large and diverse dataset for pre-training a FACS AU detector. The weights are then used to initialize the network for fine-tuning with a ``clean" dataset of manually annotated images. We show that pre-training on 100,000s of images of automatically annotated data produces a final network that achieves state-of-the-art performance, even when the network architecture is relatively simple. We release the code and data.}
    \label{fig:main}
\end{figure}

The Facial Action Coding System (FACS)~\cite{ekman1997face} is the most widely used and comprehensive taxonomy for describing facial behavior as a specific combination of building blocks known as action units. However, FACS coding is a time consuming task. Estimates put the time required to code a one-minute video at 30 minutes or more~\cite{zhao2018learning}. The demands for utilizing FACS in commercial~\cite{mcduff2016applications} and research~\cite{vijay2016computational,stratou2013automatic,kaltwang2012continuous} applications are high. Machine learning and computer vision systems are being used to automate many types of laborious labeling tasks (e.g., object and scene labeling). Facial action coding is an example of a labeling task that requires copious amounts of time and domain specific expertise. Therefore, training computer vision algorithms for automating facial action coding are very attractive. 

Automated FACS coding has a long history, comprehensive summaries of this work are available from Martinez et al.~\cite{martinez2017automatic} and Cohn and de la Torre~\cite{cohn2014automated}. 
More recently, research has been focused on using deep neural networks, specifically, convolutional neural networks (CNN), for detecting AUs~\cite{gudi2015deep, jaiswal2016deep, benitez2017recognition, niu2019local}. These methods use deep representation learning to effectively detect the presence of facial action units. However, many machine learning methods (especially deep learning approaches) are ``data hungry'', with performance monotonically increasing with the number of training samples~\cite{senechal2015facial}. Given the time consuming nature and expertise required to encode AUs, data sources that provide large numbers of training examples are limited~\cite{mavadati2013disfa, cohn2014automated, mcduff2013affectiva}. Most publicly available AU datasets have at most examples from a few hundred subjects and some only feature a few dozen. Additionally, to achieve current state-of-the-art performance, most of the published methods have made adaptations to their CNN architectures to utilize additional features for the representation learning \cite{corneanu2018deep, niu2019local, li2019self}. These modifications add complexity to model training as well as computational complexity during inference time. But are they really necessary to achieve good performance and are they the most efficient way to achieve generalization?
As network architectures and training procedures become more complex there is a growing concern about the reproducibility and replicability of machine learning findings~\cite{hutson2018artificial}. This is not helped by the lack of open data and published code, but perhaps most significantly the insufficient documentation of training parameters and environments~\cite{gunderson2018inproceedings, tatman2018practical, mcdermott2019reproducibility}. 

Recognizing these limitations, we propose a simple end-to-end pipeline to train a FACS AU classifier using a standard Convolutional Neural Network (CNN) and automatic annotations for pre-training. We show that pre-training, with these noisy labels can be beneficial and after fine-tuning with a set of clean manually labeled data, features learned can be generalizable and discriminable towards the detection of AUs (see Figure 1), even performing better than the original algorithm used to generate the noisy labels.

The concept of pre-training is widely used in machine learning to help increase generalization~\cite{erhan2010does} and has previously been applied in facial action unit classification~\cite{khorrami2015deep}. Often pre-training is performed with a proxy task like face recognition~\cite{huang2016hybrid} as available datasets for these tasks are much larger.
We show that using this simple and fundamental approach with a large set of automatically AU labeled images can perform better than the current state-of-the-art  models. In addition, we systematically looked at the impact of final AU classification results for modulating the number of images as well as the number of individuals in the pre-training stage.

In summary, this paper has the following contributions:
\begin{enumerate}
    \item To present a large set of automatically FACS-annotated images with gender, nationality and biographical meta-data.
    \item To propose a simple pipeline of pre-training and fine-tuning a CNN classifier in an end-to-end fashion for detecting the presence of facial action units that produces state-of-the-art performance.
    \item To conduct experiments to systematically investigate the effect of (1) the number of pre-training images and (2) the number of pre-training images of different people.
\end{enumerate}{}

The dataset, code, as well as relevant documentation is available for public use and can be found at:  \url{https://github.com/facialactionpretrain/facs}

\section{Related Work}

Companies now offer public software development kits (SDKs) and application programming interfaces (APIs) for FACS AU coding~\cite{mcduff2016affdex}. These computer vision techniques have been applied toward automating coding of FACS for a number of applications, see ~\cite{martinez2017automatic,takalkar2018survey} for surveys. The performance of these algorithms is highly dependent on the volume of curated training data that is available~\cite{senechal2015facial, zhu2012we}. A number of public databases are available and have been used to progress the field of automated facial action detection systems in recent years ~\cite{cohn2014automated,lucey2010extended, valstar2010induced, savran2008bosphorus, mavadati2013disfa,mcduff2018fed+}. However, many of these datasets were collected under controlled conditions with individuals performing posed expressions or have examples from a limited number of different individuals. A 2015 study~\cite{girard2015much}, showed performance of AU detection increased with a greater number of training examples from different individuals emphasizing the need for diversity in training sets.
Datasets like EmotioNet ~\cite{fabian2016emotionet} have tried to address the issue of scarcity. EmotioNet is comprised of roughly 1 Million images (950,000 labeled by algorithm and 50,000 labeled manually) labeled for 11 AUs. EmotioNet is an impressive resource and has been shown to be effective even though the accuracy of the annotations is only $\sim$80\% as reported by the authors. However, the dataset is not readily available for all researchers (including those in industry labs).

\begin{table}
    \begin{center}
            \caption{A detailed architectural description of the modified VGG13 Network we used. Output layer * denotes the output vector size for fine-tuning stage.}
        \begin{tabular}{ccc}
            \toprule
            \textbf{Type} & \textbf{Filter, stride, (drop\%)} & \textbf{Output (N, W, H)}\\ \hline \hline
            Input & - & 3, 64, 64\\ \hline
            Conv1-1/ReLu & 3x3, stride = 1 & 64, 64, 64\\ \hline
            Conv1-2/ReLu & 3x3, stride = 1 & 64, 64, 64\\ \hline
            MaxPool1/Drop1 & 2x2, stride = 2, (0.25) & 64, 32, 32\\ \hline
            Conv2-1/ReLu & 3x3, stride = 1 & 128, 32, 32\\ \hline
            Conv2-2/ReLu & 3x3, stride = 1 & 128, 32, 32\\ \hline
            MaxPool2/Drop2 & 2x2, stride = 2, (0.25) & 128, 16, 16\\ \hline
            Conv3-1/ReLu & 3x3, stride = 1 & 256, 16, 16\\ \hline
            Conv3-2/ReLu & 3x3, stride = 1 & 256, 16, 16\\ \hline
            Conv3-3/ReLu & 3x3, stride = 1 & 256, 16, 16\\ \hline
            MaxPool3/Drop3 & 2x2, stride = 2, (0.25) & 256, 8, 8\\ \hline
            Conv4-1/ReLu & 3x3, stride = 1 & 256, 8, 8\\ \hline
            Conv4-2/ReLu & 3x3, stride = 1 & 256, 8, 8\\ \hline
            Conv4-3/ReLu & 3x3, stride = 1 & 256, 8, 8\\ \hline
            MaxPool4/Drop4 & 2x2, stride = 2 (0.25) & 256, 4, 4\\ \hline
            FC5/ReLu + Drop5 & (0.5) & 1024\\ \hline
            FC6/ReLu + Drop6 & (0.5) & 1024\\ \hline
            Output/Sigmoid & - & 17(12*)\\
            \bottomrule
            \label{table:arch}
        \end{tabular}
    \end{center}
\end{table}

\begin{table}
    \centering
    \caption{The distribution of gender and geographical region (based on nationality) of the subjects in the initial pre-training dataset from the MSCeleb.}
  \begin{tabular}{ r c c }
    \toprule
     & No. Subjects & No. Images \\ \hline \hline
    Total & 1,995 & 162,070 \\ \hline
    Men & 1,070 & 82,685 \\
    Women & 925 & 79,385 \\ \hline
    N. America /\ Europe & 1,575 & 128,031 \\
    S. Asia & 97 & 9,349 \\
    S. Africa & 21 & 1,439 \\
    Mid. East /\ C. Asia /\ \\ N. Africa & 26 & 2,250\\
    Central /\ South America & 82 & 6,254 \\
    N.E Asia & 178 & 13,214 \\
    S.E Asia & 16 & 1533 \\
    \bottomrule \\
  \end{tabular}
    \label{table:data}
\end{table}

\begin{figure*}[thpb]
    \centering
    \includegraphics[width=\textwidth]{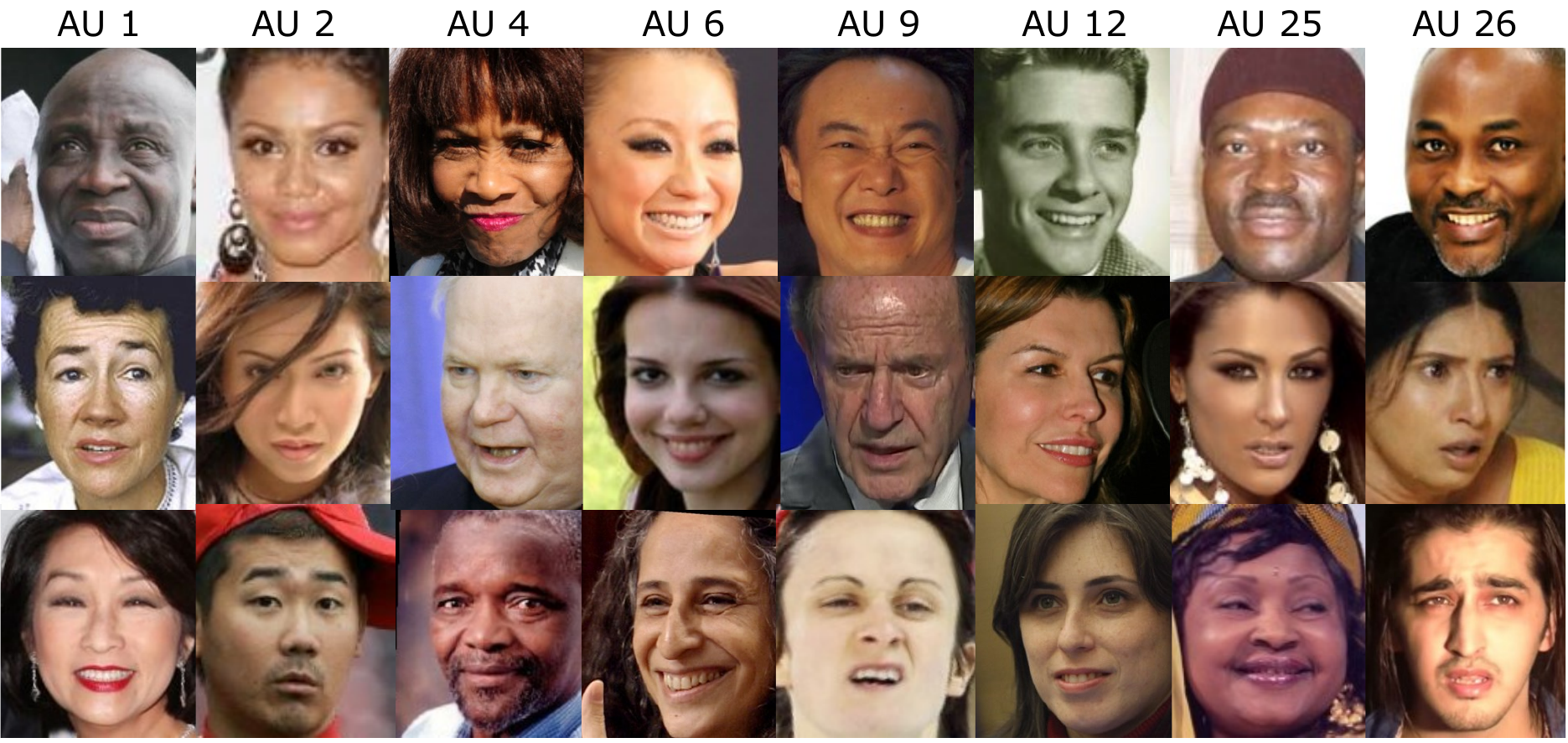}
    \caption{Images of faces labeled via OpenFace 2.0 with positive examples of each action unit (for AUs: 1, 2, 4, 6, 9, 12, 25, 26). The labels are noisy, especially in the case of co-occurring AUs. Our experiments show that pre-training of this data can still lead to improvements in performance of the final AU classifier.}
    \label{figurelabel}
\end{figure*}

Traditional feature representation-based AU detection methods focus on discriminative handcrafted features like those from Gabor wavelet~\cite{valstar2006fully} and geometry features ~\cite{baltruvsaitis2015cross}. However, the effectiveness of these features can be limited. More recently, approaches based on deep convolutional neural networks have been shown to outperform traditional approaches for AU detection~\cite{gudi2015deep,zhao2016deep, li2017action, niu2019local}. Starting from straight-forward end-to-end convolutional neural networks researchers have made adaptations to the architectures to achieve the current state-of-the-art performance. Zhao et al.~\cite{zhao2016deep} used a network (DRML) with a proposed region layer that captures local structural information in different facial regions. Li et al. \cite{li2017action} proposed using an adaptive region of interest cropping network to learn separate filters for different regions and merged them with an LSTM-based temporal fusion approach for AU detection. Shao et al.~\cite{shao2018deep} proposed an end-to-end deep learning framework for joint AU detection and face alignment. The joint learning of the two tasks and sharing features was enabled by initializing the attention maps with the face alignment results. Additionally, they also proposed an adaptive attention learning module to localize ROIs of AUs adaptively to extract better local features. Recently, Niu et al.~\cite{niu2019local} proposed an end-to-end framework which consists of a stem network for shared feature learning, a local relationship learning network, and a person-specific shape regularization network. The combination of these three modules have shown to produce state-of-the-art performance. 

Pre-training, even unsupervised pre-training, has been shown to be effective in many deep learning tasks as it supports better generalization of the resulting model~\cite{erhan2010does}. Pre-training can become particularly important in contexts where labeled training data is sparse, which is the case for facial action unit recognition~\cite{khorrami2015deep}. Inspired by this, we propose to use a large set of automatically annotated noisy AU data to pre-train a familiar and relatively simple CNN architecture, then fine-tune it with clean manually labeled data. We find that this approach can still outperform state-of-the-art models by a reasonable margin.

\begin{table*}
    \centering
    \caption{F1-frame score (in \%) as reported by LP-NET~\cite{niu2019local}, OpenFace 2.0~\cite{baltrusaitis2018openface}, and for our proposed methods on the DISFA
dataset. The best score is in bold, and bracket for the second best.}
    \begin{tabular}{l|c|c|c|c|c|c|c|c|c}
    \toprule
    \textbf{F1 Score} & \textbf{AU01} & \textbf{AU02} & \textbf{AU04} & \textbf{AU06} & \textbf{AU09} & \textbf{AU12} & \textbf{AU25} & \textbf{AU26} & \textbf{Avg}\\ \hline \hline
    OpenFace 2.0~\cite{baltrusaitis2018openface} & \textbf{41.9} & [34.7] & 66.7 & 42.6 & 36.3 & 60.8 & 90.3 & 55.8 & 53.6  \\ \hline
    LP-NET~\cite{niu2019local} & 29.9 & 24.7 & \textbf{72.7} & \textbf{46.8} & \textbf{49.6} & [72.9] & \textbf{93.8} & \textbf{65.0} & 56.9 \\ \hline
    Ours & [41.5] & \textbf{49.5} & [70.2] & [46.2] & [47.9] & \textbf{75.6} & [90.7] & [57.6] & \textbf{59.9}\\
    \bottomrule
    \end{tabular}
    \label{table:main_results}
\end{table*}


\begin{table*}
    \centering
    \caption{The F1-frame score on the DISFA dataset for each model using N number of image examples in the pre-training dataset.}
    \begin{tabular}{l|c|c|c|c|c|c|c|c|c}
    \toprule
    \textbf{N Examples} & \textbf{AU01} & \textbf{AU02} & \textbf{AU04} & \textbf{AU06} & \textbf{AU09} & \textbf{AU12} & \textbf{AU25} & \textbf{AU26} & \textbf{Avg}\\ \hline \hline
    1,000 & 0 & 0.1 & 35.4 & 33.7 & 0.4  & 69.3 & 71.8 & 13.5 & 28.0\\ \hline
    2,000 & 19.8 & 20.7 & 54.1 & 41.0 & 11.7 & 73.0 & 79.6 & 17.9 & 39.7\\ \hline
    10,000 & 8.5 & 35.9 & 61.6 & 46.7 & 34.7 & 73.7 & 83.0 & 12.9 & 44.6\\ \hline
    $\approx160,000$ & 41.5 & 49.5 & 70.2 & 46.2 & 47.9 & 75.6 & 90.7 & 57.6 & 59.9\\
    \bottomrule
    \end{tabular}
    \label{table:images}
\end{table*}

\begin{table*}
    \centering
    \caption{The F1-frame score on the DISFA dataset for each model using N number of unique people in the pre-training dataset.}
    \begin{tabular}{l|c|c|c|c|c|c|c|c|c}
    \toprule
    \textbf{N Individuals} & \textbf{AU01} & \textbf{AU02} & \textbf{AU04} & \textbf{AU06} & \textbf{AU09} & \textbf{AU12} & \textbf{AU25} & \textbf{AU26} & \textbf{Avg}\\ \hline \hline
    12 & 0 & 0 & 0 & 0 & 0  & 0 & 0 & 0 & 0\\ \hline
    200 & 19.0 & 38.8 & 64.6 & 41.3 & 23.4 & 74.9 & 84.5 & 18.8 & 45.7\\ \hline
    600 & 27.9 & 21.8 & 68.5 & 50.2 & 43.8 & 74.2 & 89.0 & 19.2 & 49.3\\ \hline
    1,000 & 48.0 & 54.6 & 70.8 & 49.2 & 45.2 & 73.9 & 88.9 & 37.4 & 58.5\\ \hline
    $\approx2,000$ & 41.5 & 49.5 & 70.2 & 46.2 & 47.9 & 75.6 & 90.7 & 57.6 & 59.9\\
    \bottomrule
    \end{tabular}
    \label{table:people}
\end{table*}

\section{Data}

It is well established that a multi-layer feed-forward network using non-linear activation function can be a universal approximator~\cite{hornik1989multilayer, leshno1993multilayer, goodfellow2016deep}. Networks with deeper architectures are beneficial for learning the kind of complicated functions that can represent high-level abstractions in vision, language, and other domains~\cite{bengio2009learning}. Training these models can be challenging since the objective function is a highly non-convex function of the parameters with a potential for having many distinct local minima in the model parameter space \cite{bengio2007scaling}. During the training process these deep models are sensitive to the initial weights and if poorly initialized, can lead to slow training, ``vanishing gradients", ``dead neurons", and/or even numerical problems. While, methods for weight initialization have been proposed to help the training process~\cite{glorot2010understanding, mishkin2015all}, they do not aim at improving generalization \cite{peng2018using}. Our proposed method is to pre-train with noisy openly available automatically annotated AU data, the learned weights are then reused as initial weights for the fine-tuning stage with clean manually labeled data.

\subsection{Pre-Training Set}

For the pre-training stage we employed the large scale publicly available MS-Celeb-1M dataset~\cite{guo2016ms}. The dataset contains over 10 million images of 1 million unique individuals retrieved from popular search engines.  We used biographical data, obtained from an Internet knowledge database, to obtain the gender and nationality of these individuals. From this dataset we then randomly sampled over 160K images for annotation to be used for pre-training our model. During the sampling an even gender split was maintained. Table~\ref{table:data} shows the distribution of gender and geographical region (based on the subject's nationality) from which the images were taken. As the distribution of subjects in the MS-Celeb dataset was heavily skewed towards those from North America and Europe, and we randomly uniformly sampled from this set, our data was also heavily skewed. Future work will consider the impact of balancing this set by region and gender, rather than just gender.

The set of images sampled from the MS-Celeb-1M dataset for pre-training was automatically annotated using OpenFace 2.0~\cite{baltrusaitis2018openface}. OpenFace 2.0 is an open-source toolkit capable of facial landmark detection, head pose estimation, facial action unit recognition, and eye-gaze estimation. OpenFace 2.0 gives estimates of both the intensity and presence of each action unit in a face image. Intensities of 17 AUs are given as a regression output from 0 to 100 while the presence of 18 AUs are given as a binary value (0 absent, 1 present). For our model development, we only focused on the presence of 17 AUs (excluding AU45). From the initial set of over 170,000 images, OpenFace 2.0 successfully annotated over 160,000. These were then used as the image set to pre-train our model. For training, we used a 95/5 training/test split. We have created a separate download link for these data with the automatic annotations. The data can be found at the URL: \url{https://github.com/facialactionpretrain/facs}

\subsection{Finetuning and Testing}

For fine-tuning our model we employed the DISFA dataset~\cite{mavadati2013disfa}. The dataset contains videos of 27 young adults (15 males and 12 females) who were asked to watch a 4-minute video clip intended to elicit spontaneous expressions of emotion. Each video frame was manually coded for the presence, absence, and intensity levels (0 to 5) of 12 AUs. For our experiment, frames with intensity levels equal or greater than 2 were labeled as positive examples and the rest are labeled as negative examples. About 130,000 frames were used in the final experiments. 

\section{Proposed Method and Experimental Setup}

Table~\ref{table:arch} shows a detailed overview of the modified VGG13 architecture used for our facial AU detection model. The model contains convolutional layers with max pooling.  We used dropout to help avoid overfitting~\cite{srivastava2014dropout}. The convolutional layers were followed by two fully connected layers. As described below the final fully connected layer was replaced between the pre-training and fine-tuning steps.

\subsection{Pre-trainining}
For the pre-training stage, we used an Atom Optimizer with a learning rate of 0.005, momentum rate of 0.9 and a batch size of 32. The network was allowed to train until either convergence or a maximum epoch of 500 was reached.

\subsection{Fine-Tuning}
For the fine-tuning stage, we replaced the final fully connected (fc) network output layer from out pre-trained model with a new output layer mapping to the 12 annotated AUs in the DISFA dataset. For training, a subject independent 3-fold cross-validation protocol was used. The network was optimized with an Atom Optimizer with a learning rate of 0.0001, a momentum rate of 0.9 and a batch size of 32. The network was allowed to train and we employed an early stoppage criteria of 10 epochs.
Both the pre-training and fine-tuning network used a binary cross entropy loss function. All implementations were created using CNTK~\cite{seide2016cntk}. 

\subsection{Image Pre-processing}
\subsubsection{Pre-Training}
All input images were mean-normalized, converted into a single channel grayscale format and resized to 64x64. Random horizontal flip, rotation, skew, and scale were used for data augmentation. Image resizing and data augmentation was performed online during training.

\begin{figure}
    \centering
    \includegraphics[width=3.4in]{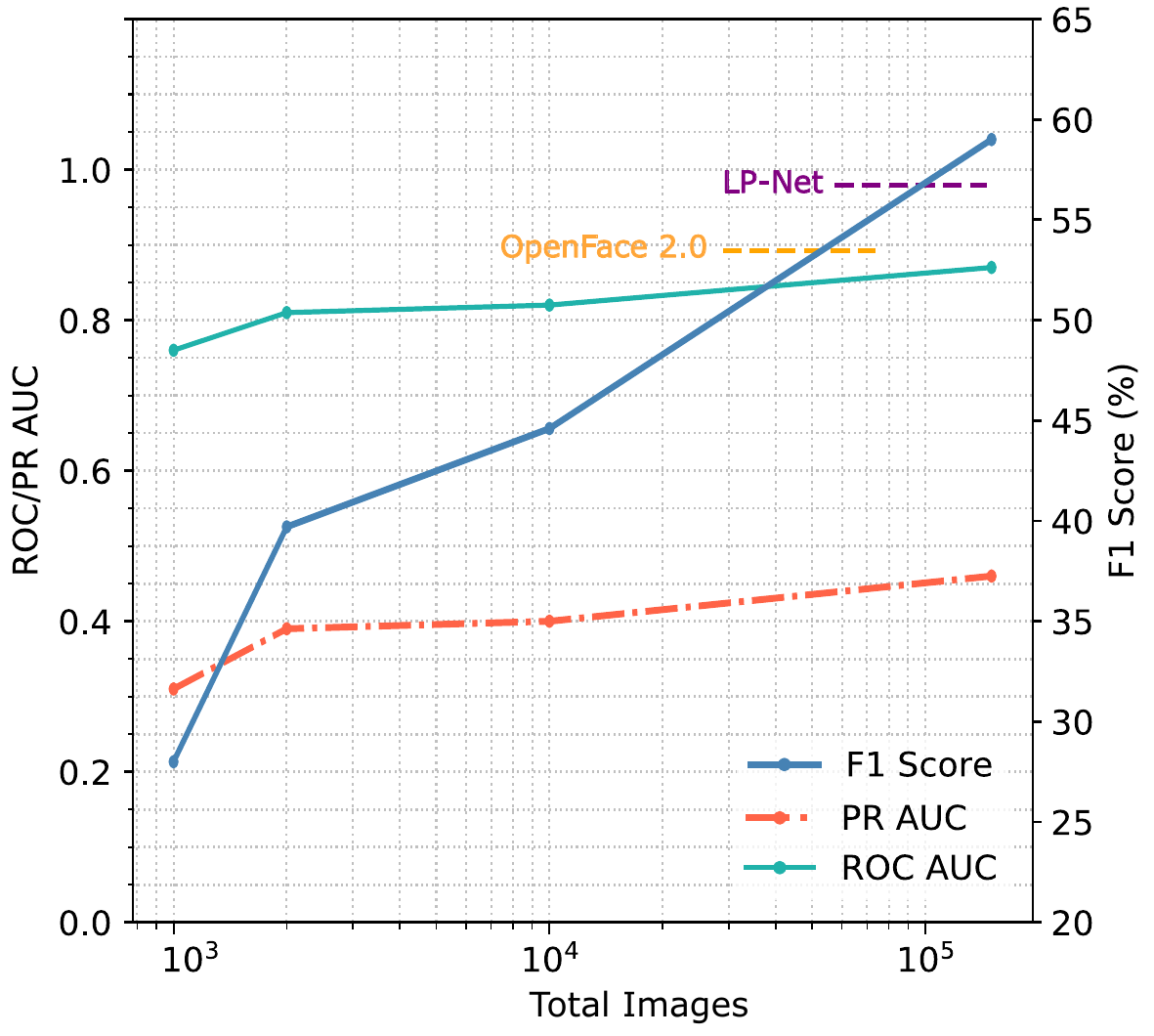}
    \caption{ROC AUC, PR AUC and F1 scores of our model as a function of the number of automatically annotated images used at the pre-training step. The performance of LP-Net and OpenFace 2.0 are shown for comparison.}
    \label{fig:Images}
\end{figure}

\begin{figure}
    \centering
    \includegraphics[width=3.4in]{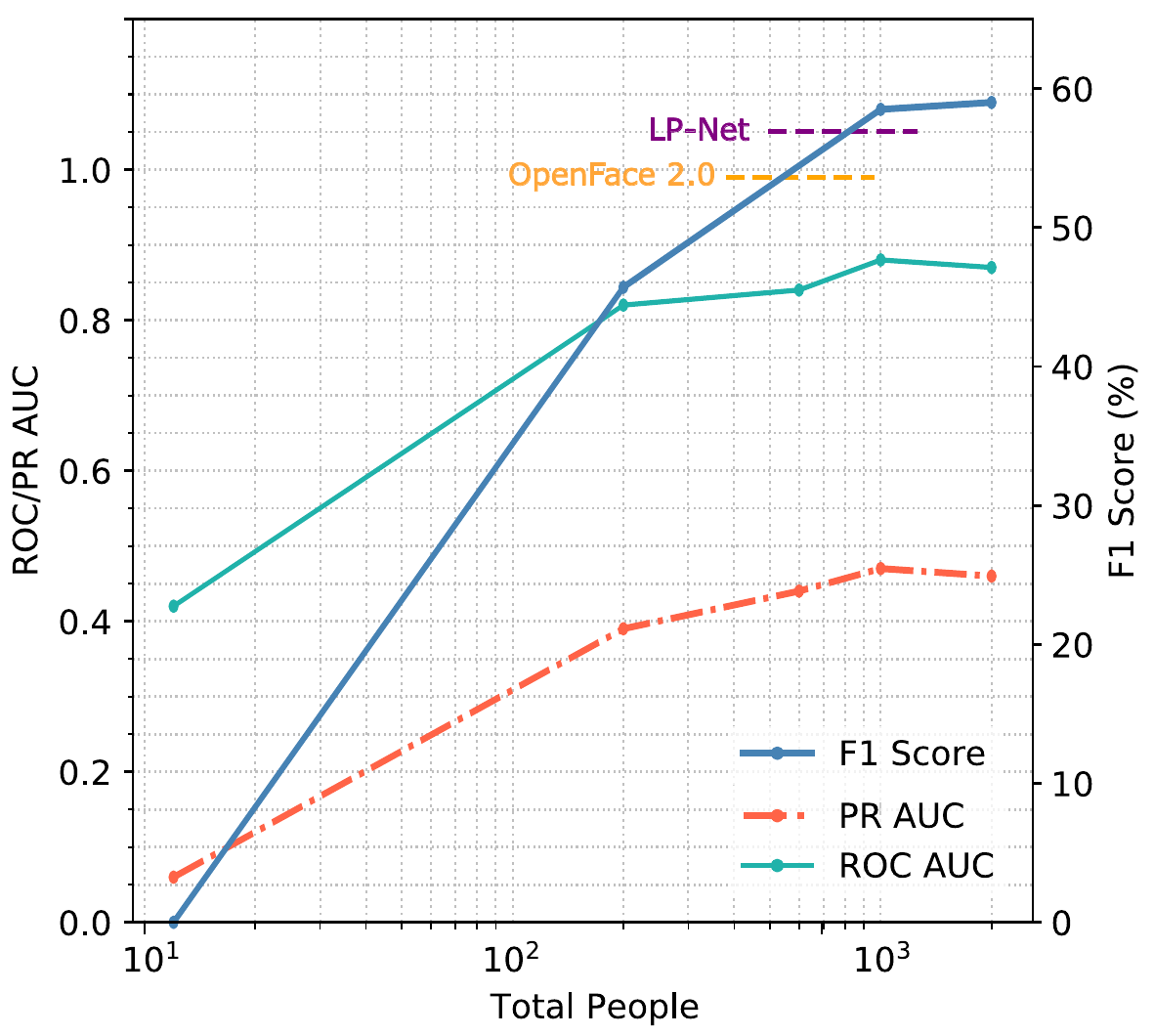}
    \caption{ROC AUC, PR AUC and F1 scores of our model as a function of the number of automatically annotated images \emph{of different people} used at the pre-training step. The performance of LP-Net and OpenFace 2.0 are shown for comparison.}
    \label{fig:People}
\end{figure}

\subsubsection{Fine-Tuning}
OpenCV's ~\cite{opencv_library} deep neural network (DNN) face detector was used to locate and crop out the faces for each of the frames in the DISFA dataset. The cropped images were then zero padded to maintain a square (1:1) aspect ratio, mean-normalized, converted into a single channel grayscale format and resized to 64x64. Random horizontal flip, rotation, skew, and scale were used for data augmentation. As with the pre-training, image resizing and data augmentation was performed online during training.

\subsection{Evaluation Metrics}
For comparison with other methods, performance was evaluated using the F1-score. The F1-score is the harmonic mean of precision and recall and was calculated by binarizing the output results at a threshold of 0.5. We also computed the areas under the precision recall (PR) curve, receiver operating characteristic (ROC) curve to capture the overall model performance and not just at a specific operating point. All metrics were computed per AU and also then averaged across AUs. 

\subsection{Varying Number of Images \& People during Pre-Training}
Additional experiments were conducting to further understand the effects of independently varying the number of images as well as the number of individuals in the images during the pre-training stage. From the set of 162,070 OpenFace 2.0 annotated images, subsets of randomly sampled images containing 1,000, 2,000 and 10,000 face images were created, the images were uniformly sampled not considering the individuals within the face images. Then another set of data subsets were created with images specifically sampled from each of 12, 200, 600 and 1000 different people.  These subsets were used to pre-train different VGG-13 models. As in the original 162,070 images, a gender balanced split was maintained for each of these image subsets. Identical fine-tuning procedures were performed on the different pre-trained models using the DISFA dataset and the subject independent 3-fold cross-validation scheme described above.

\section{Results}
\subsection{Comparison with State-of-the-art}

The results of our model were compared against the current state-of-the-art LP-Net~\cite{niu2019local} and OpenFace 2.0~\cite{baltrusaitis2018openface} using the F1-score metric (as this is the consistent metric used across all the papers). To maintain consistency with the authors of LP-Net, 8 of the 12 AUs were used for the comparison. Our approach achieves the best performance in terms of overall average F1-score and achieves either the best or second best performance for the AUs annotated in DISFA. The overall average F1-score metrics, PR area, and ROC area achieved on the DISFA dataset with our model was 0.60, 0.45, and 0.86 respectively. Table~\ref{table:main_results} shows the results of the comparisons on the DISFA dataset. These results suggest that our method has good generalizability and can achieve state-of-the-art performance even when compared to methods using more complex network architectures.  The performance improvement on AU01 and AU02 was particularly large using the pre-training approach.  The F1-score for AU02 was over 100\% greater compared to LP-Net~\cite{niu2019local}. Perhaps this is due to the fact that the pre-training allowed the algorithm to ``see" more examples of people with different facial appearances.

\subsection{The Effect of Number of Examples on Performance}

Results from our systematic experiments designed to understand the effects of varying the number of images and individuals can be seen in Figures~\ref{fig:Images} and \ref{fig:People}. Numerical results can be seen in Tables~\ref{table:images} and \ref{table:people}, respectively.

Specifically, Figure~\ref{fig:Images} shows the F1, ROC and PR metrics when pre-training with different number of total image examples (with equal numbers of men and women). The F1 score performance of LP-Net and OpenFace 2.0 are shown for reference. The performance metrics for LP-Net are reported in ~\cite{niu2019local} while the F1 score performance of the OpenFace 2.0 classifier was calculated on the DISFA dataset to provide the appropriate comparison. The results show that as you increase the number of images performance increases monotonically in a close to linear fashion.

\subsection{The Effect of Number of Subjects on Performance}

Figure~\ref{fig:People} shows the F1, ROC and PR metrics when pre-training with images of different numbers of people (with equal numbers of men and women). The F1 score performance of LP-Net and OpenFace 2.0 are shown for reference. The performance metrics for LP-Net are reported in ~\cite{niu2019local} while the F1 score performance of the OpenFace 2.0 classifier was calculated on the DISFA dataset to provide the appropriate comparison. Again, the results show that as you increase the number of images of different people performance also increases monotonically, initially at a steeper rate than for the previous plot. This suggests that more images of different people is more beneficial than more images of the same people.

\section{Discussion and Conclusion}

Facial action coding is an important tool in facial analysis and affective computing. FACS provides a useful and objective coding mechanism. However, coding images and videos is extremely labor intensive and requires a level of training that many researchers may not have. Machine learning techniques are being successfully utilized for FACS recognition. We show that pre-training on a diverse but noisy set of images can lead to simple network architectures out-performing more complex architectures and obtaining state-of-the-art results. We find that when the labels used for pre-training are generated with an existing set of AU detectors (in this case OpenFace 2.0) the final model even outperforms the original detectors.  This suggests that the model is able to learn to generalize from these noisy data and form representations that are more effective. This is most likely due to the fact that the model is able to further separate the signal (AU appearances) from noise (different face shapes, head poses and other variations) by observing a very large number of different faces. Our results are supported by the fact that when we experiment with different numbers of unique individuals in the pre-training set (see Figure~\ref{fig:People}) the performance increases dramatically with the diversity - even though the label quality, from the automated algorithm - OpenFace 2.0 - is still imperfect. One could describe this as a form of semi-supervised learning. It suggests that increasing the diversity (number of subjects) in a pre-training set, but losing some accuracy in the labels, is still beneficial compared to training only on a smaller set of more accurately labeled images. The process of creating this pre-training data is very scalable and we release our set with this paper.

Our results are in agreement with Girard et al.~\cite{girard2015much} who showed performance for an AU classifier trained on appearance based features greatly increased as the number of subjects in the training set increased. In their experiment increasing the training data with manually labeled AUs from 8 to 64 subjects resulted in a significant increase in classification performance. This further supports our hypothesis that pre-training from a dataset containing images of a very large number of unique individuals can help the model learn representations that separate signal from noise even when the signal quality is noisy. 

The approach that we have outlined in this paper uses an open-source automated FACS annotation tool and face images scraped from the Internet. As such, it is a highly scalable method. While we showed the efficacy of this approach with a simple CNN, it is in theory network architecture independent. We hope that by releasing these data and code that other researchers can leverage the benefits of noisy pre-training. 

As a proof-of-concept, we only used about 160,000 images from the MS-Celeb dataset for this work.  However, the dataset contains a full 10 million images from 1 million individuals. The pre-training dataset could be extended and our results suggest that there may still be further performance gains that can be achieved by simply increasing the size of the pre-training set (see Figure~\ref{fig:Images}).
 
\section{Distribution}
We release the dataset use in this paper alongside the code. The dataset may be used for academic and commercial research purposes. The license details, the permissible use of the data and the appropriate citation, can be found at: \url{https://github.com/facialactionpretrain/facs}. The dataset is available for distribution to researchers online.  

A summary of the dataset is included below:

\textbf{Images.} 162,070 RGB images of aligned and cropped faces of 1,995 subjects. The images are of celebrities that allows us to put biographical data as described below.

\textbf{Biography.} We searched the Google Knowledge database for the name of each celebrity in the images. This biographical text is included in the dataset in JSON format. 

\textbf{Gender.} We used pronoun counts to infer gender from the biographical data and the text using NLTK \cite{birdNaturalLanguageProcessing2009} and named-entity analysis to extract the gender. 

\textbf{Nationality/Country of Origin.} We queried the biographical data for each subject using NLTK and named-entity analysis to extract the nationality.


\balance{}

\bibliographystyle{ieee}
\bibliography{sample_FG2019}

\addtolength{\textheight}{-3cm}   

\end{document}